**Evaluating and Enhancing Large Language Models' Performance in Domain-specific Medicine: Osteoarthritis Management with DocOA**

Xi Chen[1,2,3] M.D., MingKe You[1,2] M.Sc., Li Wang[3] Ph.D.; WeiZhi Liu[3] Ph.D.; Yu Fu[3] Ph.D.; Jie Xu[5] Ph.D.; Shaoting Zhang[5] Ph.D.; Gang Chen[1,2] M.D., Ph.D.; Kang Li[3,4,5] *Ph.D.; Jian Li[1,2]*M.D.

[1]Sports Medicine Center, West China Hospital, West Chian School of Medicine, Sichuan University, Chengdu, Sichuan, China

[2]Department of Orthopedics and Orthopedic Research Institute, West China Hospital，Sichuan University, Chengdu, Sichuan, China

[3]West China Biomedical Big Data Center, West China Hospital, Sichuan University, Chengdu, Sichuan,610041, China

[4]Med-X Center for Informatics, Sichuan University, Chengdu, Sichuan, 610041, China

[5]Shanghai Artificial Intelligence Laboratory, OpenMedLab, Shanghai, 200030, China.

Corresponding Author(s): likang@wchscu.cn; lijian_sportsmed@163.com

Contributing Author(s): geteff@wchscu.edu.cn; 1217696738@qq.com; wangli1@stu.scu.edu.cn; 2020141410222@stu.scu.edu.cn; fy020617@163.com; xujie@pjlab.org.cn; zhangshaoting@pjlab.org.cn; drchengang@hotmail.com.

**Evaluating and Enhancing Large Language Models' Performance in Domain-specific Medicine: Osteoarthritis Management with DocOA**

**Evidence before this study**

General medical benchmarks and training data for LLMs often lacked depth in domain-specific content and there was a recognized need for LLMs to bridge the gap between theoretical knowledge and practical clinical application.

Before this study, there was no established benchmark framework dedicated to assessing the clinical capabilities of LLMs in specific medical domains.

While there have been efforts to develop medical LLMs, the use of retrieval-augmented generation (RAG) and instruction prompts to enhance their domain-specific clinical abilities has not been reported.

**Added value of this study**

This study developed a novel, multitiered benchmark framework for testing LLMs in specific disease management, such as osteoarthritis. This framework can be adopted in other specific medical scenarios as well.

This study integrated retrieval-augmented generation (RAG) and instruction prompts with GPT-4 to create a specialized assistant, DocOA, which significantly enhanced its abilities. This approach offers a cost-effective method for developing domain-specific medical LLMs."

This study provided a comprehensive evaluation combining objective benchmarks with human evaluations, which revealed limitations in the clinical capabilities of general LLMs.

**Implications of all the available evidence**

Our results underscore the importance of domain-specific benchmarks for precisely assessing the clinical effectiveness of large language models (LLMs). It also notes the effectiveness of using retrieval-augmented generation (RAG) and instruction prompts in enhancing the domain-specific performance of LLMs. This approach may be recognized as a valid and cost-effective strategy for creating specialized medical LLMs.

**Abstract**

The efficacy of large language models (LLMs) in domain-specific medicine, particularly for managing complex diseases such as osteoarthritis (OA), remains largely unexplored. This study focused on evaluating and enhancing the clinical capabilities of LLMs in specific domains, using osteoarthritis (OA) management as a case study. A domain specific benchmark framework was developed, which evaluate LLMs across a spectrum from domain-specific knowledge to clinical applications in real-world clinical scenarios. DocOA, a specialized LLM tailored for OA management that integrates retrieval-augmented generation (RAG) and instruction prompts, was developed. The study compared the performance of GPT-3.5, GPT-4, and a specialized assistant, DocOA, using objective and human evaluations. Results showed that general LLMs like GPT-3.5 and GPT-4 were less effective in the specialized domain of OA management, particularly in providing personalized treatment recommendations. However, DocOA showed significant improvements. This study introduces a novel benchmark framework which assesses the domain-specific abilities of LLMs in multiple aspects, highlights the limitations of generalized LLMs in clinical contexts, and demonstrates the potential of tailored approaches for developing domain-specific medical LLMs.

**Introduction**

The rapid development of large language models (LLMs) has shown promising potential in the medical field, as demonstrated by their ability to pass the United States Medical Licensing Examination (USMLE) and diagnose clinical conditions[1-3]. The promising performance of LLMs in the general medical field warrants further research and exploration of their clinical performance in domain-specific medical scenarios[4,5].

Osteoarthritis (OA) is one of the most prevalent and debilitating diseases that causes pain, disability, and loss of function[6]. The global prevalence of OA is approximately 7.6% (595 million people), as of 2020 [7]. The management of OA requires complex strategies that encompass a variety of pharmacological treatments, lifestyle alterations, rehabilitation, and surgical interventions across multiple disciplines. Effective management of this condition necessitates the integration of extensive evidence-based medical data and the consideration of individual circumstances[6].

Although some LLMs have achieved commendable results in general medical question-answering (QA) tasks, substantial limitations persist in their clinical capability, particularly in complex and multifaceted diseases such as OA[8]. However, the datasets used to train LLMs are predominantly composed of general medical knowledge and lack in-depth, domain-specific content. Existing research indicates that current training data and benchmarking methodologies may be inadequate for LLMs to acquire the necessary domain-specific knowledge and clinical capabilities[5].

Additionally, LLMs may lack the ability to translate their knowledge into clinical proficiency. Despite possessing sound knowledge about certain diseases, effectively applying this knowledge to disease diagnosis remain challenging for LLMs[9]. This observation highlights the need to train and evaluate LLMs using datasets that are more closely aligned with clinical applications, thereby bridging the gap between theoretical knowledge and practical clinical usage.

To address these challenges, we proposed to build a dataset that focuses on specific medical diseases, which should encompass updated evidence-based medical knowledge capable of providing both physicians and patients with expert disease-related information. In addition, real-world cases featuring patient information and treatment decisions encountered in clinical practice should be included. This repository can serve as a benchmark for testing the performance of LLMs in specific medical domains, such as OA management.

In addition, the integration of retrieval-augmented generation (RAG) and instruction prompts enabled the model to assimilate external knowledge and adhere to instructions in a predetermined manner. Instruction prompts facilitated the model to emulate the perspective of a digital doctor by offering professional medical advice while maintaining response consistency and avoiding hallucinations. RAG is an artificial intelligence (AI) framework that improves LLMs by integrating relevant information from external knowledge bases, thereby enhancing the accuracy and reliability of the model's responses and providing efficient and cost-effective access to updated external data[10].

In general, we propose a dataset framework that encompasses updated evidence-based medical knowledge, and real-world cases may effectively examine the capabilities of LLMs in clinical practice. The integration of RAG and prompt engineering may allow trained LLMs such as GPT-4 to acquire domain-specific abilities. Moreover, the management of OA serves as an ideal example in terms of its clinical significance and data volume on this research topic. Therefore, this study aimed to curate a dataset for OA management, evaluate knowledge of updated evidence-based medicine for LLMs and their capabilities in clinical scenarios, and adopt RAG and instruction prompts to enhance these capabilities.

**Methods**

This study curated an OA management dataset based on clinical guidelines and real-world cases. A benchmark was developed to evaluate the clinical knowledge and capabilities of LLMs for OA management. DocOA was built with instruction prompts and RAG and was tested along with other LLMs. Figure 1 illustrates the flow diagram of the study.

Figure 1. Study Flow diagram

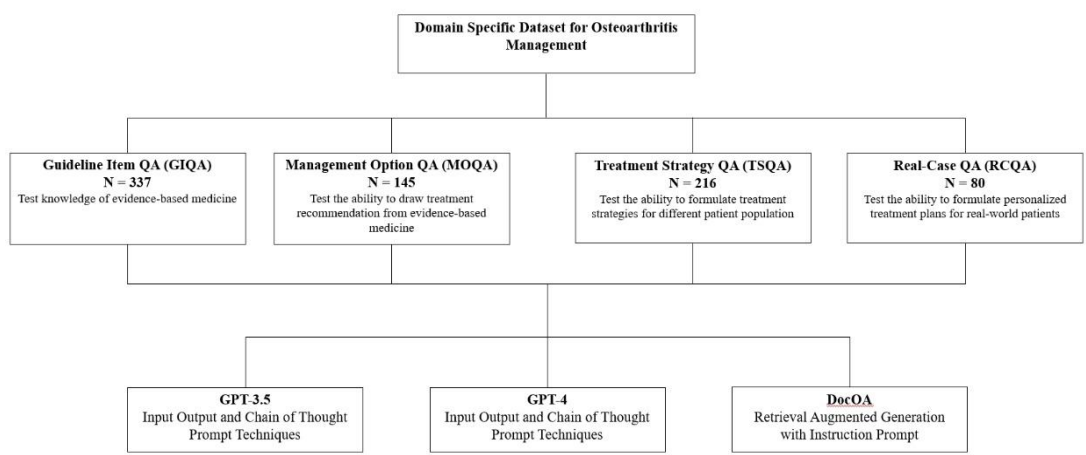

**Dataset**

This dataset was developed based on key clinical guidelines and real-world patients. After the panel discussion, six well-acknowledged guidelines and data from 80 real-world patients were selected that included various aspects of OA management. The following guidelines were included: American Academy of Orthopedic Surgeons (AAOS) management of osteoarthritis of the knee (Non-arthroplasty)[11]; National Institute for Health and Care Excellence (NICE) guideline for osteoarthritis in over 16s[12]; Osteoarthritis Research Society International (OARSI) guidelines for the non-surgical management of knee, hip, and polyarticular osteoarthritis[13]; Royal Australian College of General Practitioners Guideline for the management of knee and hip osteoarthritis[14]; American College of Rheumatology/Arthritis Foundation (ACR) Guideline for the Management of Osteoarthritis of the Hand, Hip, and Knee[15]; European League Against Rheumatism (EULAR) recommendations for the non-pharmacological core management of hip and knee osteoarthritis[16]. Between 1 April 2023 to 1 October 2023 80 patients diagnosed with osteoarthritis and who had received OA management at our hospital were randomly selected. The patient information, including age, sex, height, weight, body mass index (BMI), laterality of knee involvement, medical history, level of pain, mechanical symptoms, physical examination results, and radiographic findings, were retrieved. All identifiable information was concealed to maintain confidentiality.

The OA benchmark aims to test the clinical capabilities of LLMs at four levels within the context of evidence-based medicine, ranging from domain-specific knowledge to clinical capabilities. The benchmark assesses the performance of LLMs pertaining to OA knowledge, summarising the knowledge to formulate recommendations for specific management options, providing tailored management options for different patient populations, and formulating personalised management plans for real-world cases.

**Assistant with RAG and Instruction Prompting**

DocOA, a specialised assistant, was developed based on the GPT-4-1106-preview model, which integrates instruction prompts and RAG to enhance performance. The instruction prompt emphasised its role in providing evidence-based medical insights and personalised management programmes guided by evidence-based medicine. The DocOA strictly adheres to facts, avoids speculation, and clearly states its limitations. Moreover, it maintains a professional and informative tone suitable for medical discussions.

RAG has been used to respond to various OA-related queries. The RAG integrates a model's language generation capabilities with a retrieval system, enabling access to specific information from external sources[17]. Of the several RAG techniques and data structures tested, the retrieval function from OpenAI was adopted, and the most optimal data structure was selected and converted into the JavaScript Object Notation (JSON) format for optimal retrieval accuracy. In response to OA-related queries, the RAG enables the assistant to dynamically pull relevant data from the external dataset as it generates responses. The workflow of the assistant is illustrated in Figure 2.

Figure 2 Workflow of DocOA

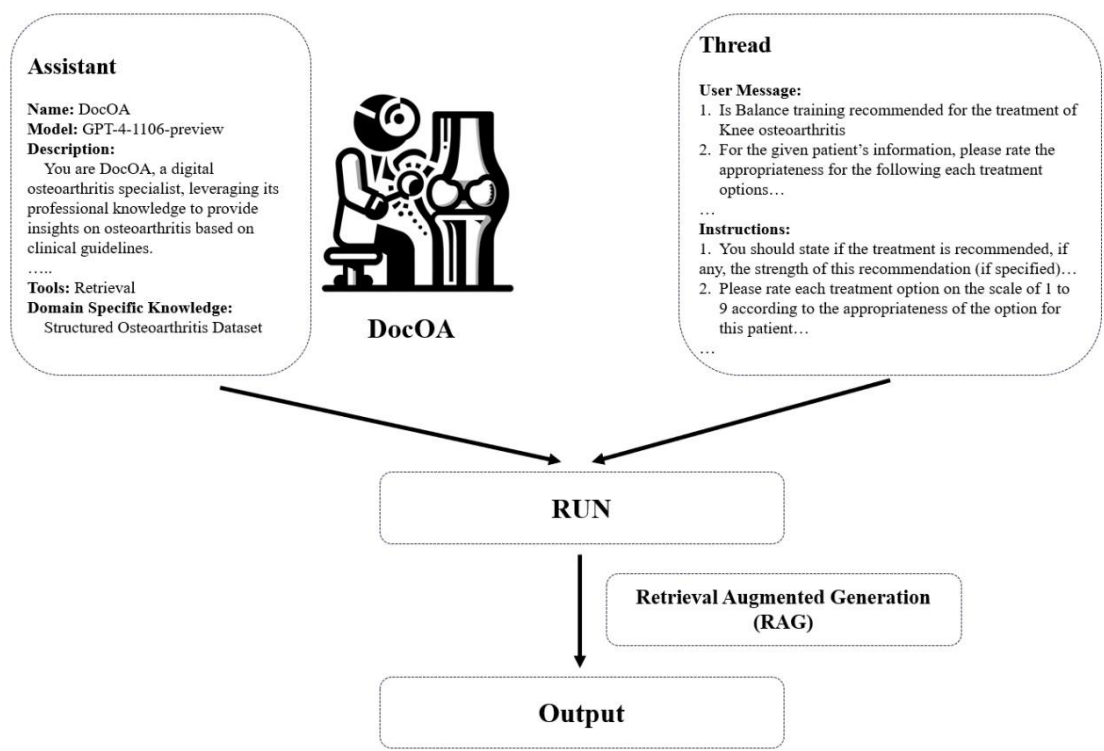

The assistant block details the core of the DocOA system, utilising a base model of GPT-4-1106-preview. The description within the block serves as an instruction prompt that outlines the system's role. The Assistant's functionality includes a Retrieval tool that accesses external dataset.

The system receives input from thread blocks in the form of questions about osteoarthritis management and requests evaluation. This includes user messages with specific queries regarding osteoarthritis treatment and detailed instructions for the system to follow.

The execution command then triggers the processing of input data through the DocOA system. Through retrieval-augmented generation (RAG), DocOA adopts external information from the knowledge database and adheres to the instructions to generate the final output.

**Models Testing**

DocOA and the two base models, GPT-3.5 and GPT-4, were tested against the OA benchmark. Each question was presented five times to each model to assess the robustness of its performance. Additionally, the zero-shot chain of thoughts (COT) prompt technique was tested for GPT-3.5 and GPT-4 to determine whether it outperformed the input-output (IO) technique.

**Evaluation of LLMs' Performance**
**Objective Evaluation**

The model-generated responses were compared with predefined correct answers for each subset of the benchmark. An answer was considered accurate if LLM provided correct knowledge (recommendation status and recommendation strength) about the treatment option and predicted the correct treatment recommendation (treatment appropriateness) for a specific patient profile or individual patient.

**Human Evaluation Framework**

The human evaluation framework is an effective approach for identifying the gap between LLMs and clinical experts[3]. In this study, human evaluation was performed by both physicians and patients. Eighty items from the OA benchmark were randomly selected for detailed human evaluation framework.

A panel of five physicians, each with a minimum of 10 years of experience in OA management, conducted the physician evaluation. The sequence of answers was randomised and the generating models were anonymized to ensure that the evaluation was conducted without any knowledge of the model that generated them. The evaluation metrics were established based on a previous study with modifications[3]. The physician assessed the quality of the responses in the following domains: inaccurate content, relevance, hallucinations, missing content, likelihood of possible harm, extent of possible harm, and possibility of bias. The ability of LLM to achieve correct comprehension, retrieval, and reasoning was assessed using the method described in a previous study[18]. Patient evaluation was conducted by assessing the user intent fulfilment, and helpfulness of the content. The detailed descriptions of each human evaluation metric are provided in Supplementary File 1.

**Statistical Analysis**

All statistical analyses were performed using the SPSS 25.0 software (IBM, Armonk, NY, USA) and GraphPad Prism version 8 (GraphPad Software, San Diego, CA, USA). Discontinuous data are expressed as incidence and rate and analysed using the chi-square test for differences. A P value less than 0.05 indicated statistical significance.

## Results

**OA Benchmark**

The benchmark comprised four subsets of question-answer (QA) evaluations designed to test the performance of LLMs across a spectrum ranging from domain-specific knowledge to practical capability. Guideline-item QA (GIQA), which was developed based on specific items extracted from the clinical guidelines, evaluates the LLMs' knowledge of these well-established standards. The GIQA comprised 337 items. Management options QA (MOQA) included summarised recommendations for specific treatments from the included clinical guidelines. The MOQA, which comprised 145 items, evaluated LLMs' knowledge of specific treatment options, as well as their ability to summarise medical evidence. Treatment strategy QA (TSQA), which included treatment recommendations for different patient populations, provided treatment recommendations based on the patient's age, clinical presentation, and other factors. The TSQA, which comprised 216 items, evaluated the capability of LLMs to derive treatment recommendations for specific patient types. Real-case QA (RCQA) included treatment recommendations for 80 real-world patients. The RCQA, which comprised 80 items, evaluated LLMs' capability in formulating treatment recommendations in a more complicated scenario in which individual information is provided, mirroring real-world clinical decision-making. This dataset is available on GitHub. Examples of each QA type are shown in Figure 3.

Figure 3 Benchmark framework for osteoarthritis management

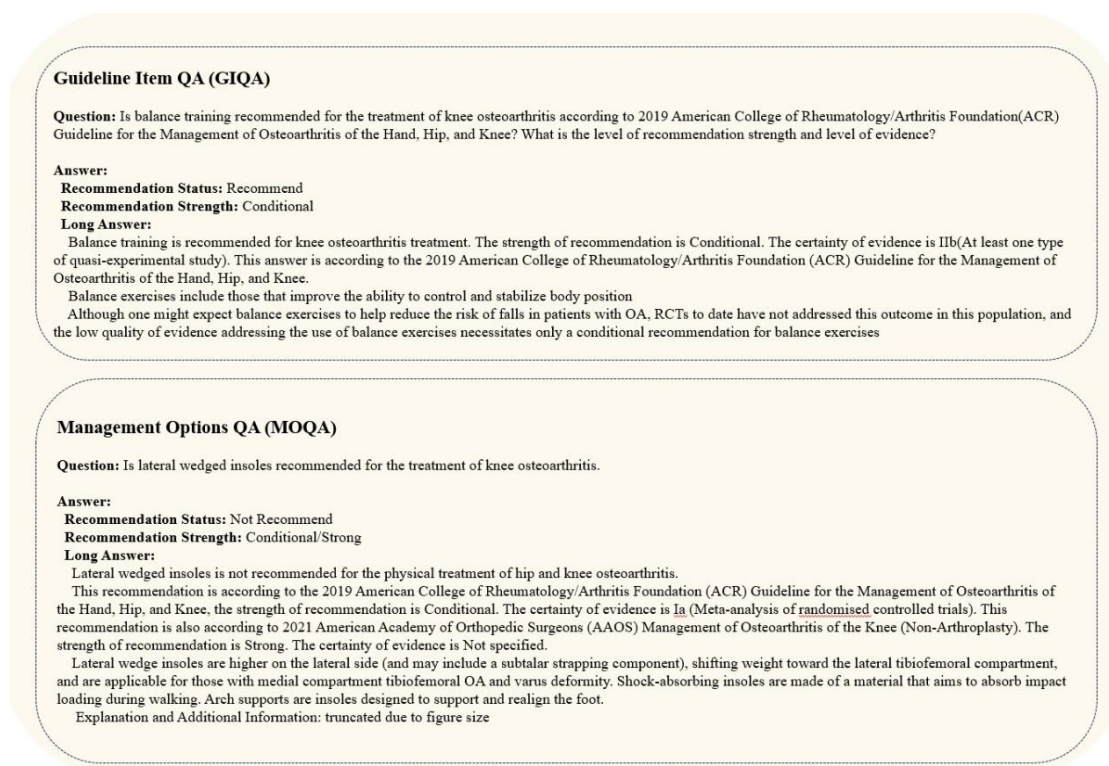

**Treatment Strategy QA (TSQA)**

**Question:**
For the given patient profile, please rate the appropriateness for each treatment option. The patients is middle-aged; The patient has function-limiting pain that is intermittent and predictable; The patient has arthritic involvement predominantly in one weight bearing compartment. Radiographs show minimal joint space narrowing (KL 0-1). In terms of mechanical symptoms suggesting loose body or meniscal tear, the patient has mechanical symptoms present.
The treatment options to be evaluated include: **1.** Self management program; **2.** Prescribed Physical Therapy; **3.** Hinged Knee Brace and/or Unloading Brace, Assistive Devices; **4.** NSAID or Acetaminophen; **5.** Arthroscopic Partial Meniscectomy or Shaving; **6.** Intraarticular Corticosteroids; **7.** Platelet rich plasma (PRP).

**Answer:**
The appropriateness of each treatment option for this type of patient profile is: **1.** Self-Management Programs (unsupervised exercise, tai chi, weight loss, aerobic walking) is Appropriate Rating: 8; **2.** Prescribed Physical Therapy (Supervised Exercise, manual therapy, neuromuscular training, etc.) is Appropriate Rating: 8; **3.** NSAID or Acetaminophen is Appropriate Rating: 8; **4.** Intraarticular Corticosteroids is Appropriate Rating: 8; **5.** Arthroscopic Partial Meniscectomy or Shaving is Appropriate Rating: 7; **6.** Hinged Knee Brace and/or Unloading Brace, Assistive Devices (e.g., cane, walker) is May Be Appropriate Rating: 6; **7.** PRP is Rarely Appropriate Rating: 3

**Real-Case QA (RCQA)**

**Question:**
For the given patient's condition, please rate the appropriateness of the following treatment options. The patient is a 37 years old female who is experiencing discomfort in the bilateral knee. The patient is 165 cm in height and 60 kg in weight, the BMI is 22.03 kg/m2. The patient's past medical history is allergic to NSAIDs. The patients has function-limiting pain that is constant with intense intermittent unpredictable episodes. The patient has arthritic Involvement predominantly in one weight bearing compartment. Imaging and physical examination revealed severe joint space narrowing (KL 4) with significantly limited range of motion. The patient has mechanical symptoms that is compatible with meniscal tear or loose body.

**Answer:**
For this patient, the appropriateness of each treatment option for this patient is as follows: **1.** Self-management programs (unsupervised exercise, tai chi, weight loss, aerobic walking) is Appropriate and the rating is 8. **2.** Prescribed physical therapy (supervised exercise, manual therapy, neuromuscular training, etc.) is Appropriate and the rating is 8. **3.** Hinged knee brace and/or unloading brace, assistive devices (e.g., cane, walker) is Appropriate and the rating is 7. **4.** NSAID or Acetaminophen is Not appropriate. **5.** Intraarticular corticosteroids is Appropriate and the rating is 9. **6.** Arthroscopic partial meniscectomy or shaving is Rarely appropriate and the rating is 3. **7.** Platelet rich plasma is Rarely appropriate and the rating is 3.

**Objective Evaluation**

The accuracy of GPT-3.5 in GIQA, MOQA, TSQA and RCQA was 0.26, 0.22, 0.01 and 0.03, respectively. The accuracy of GPT-4 in GIQA, MOQA, TSQA, and RCQA was 0.38, 0.30, 0.07, and 0.01, respectively. The accuracy of DocOA in GIQA, MOQA, TSQA, and RCQA was 0.92, 0.87, 0.88, and 0.72, respectively. The accuracy of each model against the benchmark is presented in Table 1 and Figure 4a. As shown in Figure 4a, the degree of accuracy significantly decreased: GIQA > MOQA > TSQA > RCQA. As shown in Figure 4b, DocOA reported 111 failures in accessing the external dataset, which accounted for 12.4% of the inaccurate answers generated.

Figure 4 Results of objective evaluation

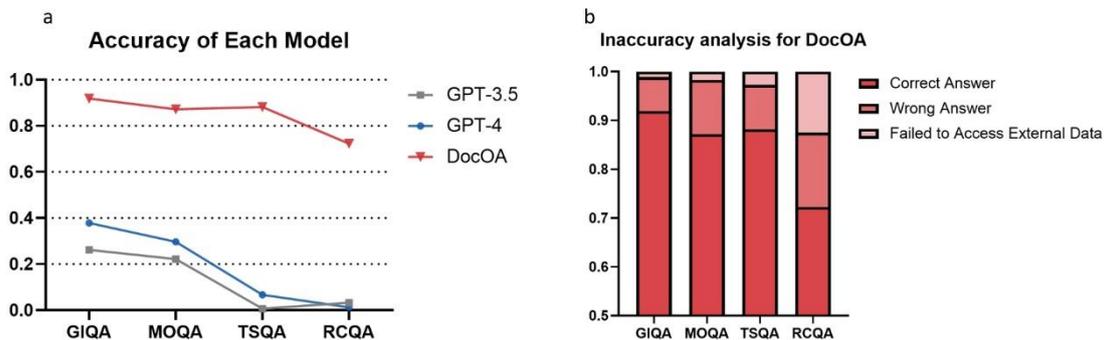

a. accuracy of each model against each subset of benchmark; b. Inaccuracy analysis for DocOA due to wrong answer and failure in access external data.

Table 1. Accuracy of each model against osteoarthritis benchmark

|  | GPT-3.5 | GPT-4 | DocOA | P Value |
|---|---|---|---|---|
| Osteoarthritis Benchmark | 0.16 | 0.24 | 0.88 | <0.001* |
| Guideline item QA (GIQA) | 0.26 | 0.38 | 0.92 | <0.001* |
| Management option QA (MOQA) | 0.22 | 0.30 | 0.87 | <0.001* |
| Treatment strategy QA (TSQA) | 0.01 | 0.07 | 0.88 | <0.001* |
| Real-case QA (RCQA) | 0.03 | 0.01 | 0.72 | <0.001# |

*Further analysis showed all pairwise comparison had P value less than 0.05; #Further analysis showed P value is 0.056 for GPT-3.5 vs GPT-4, <0.001 for GPT-3.5 vs DocOA, <0.001 for GPT-4 vs DocOA.

Zero-shot chain of thoughts (COT) prompt techniques were adopted for GPT-3.5 and GPT-4. Compared with the input-output (I/O) prompt technique, no significant improvements in model performance were observed. The results are summarised in Table 2.

Table 2. Accuracy of different prompt techniques against osteoarthritis benchmark

|  | GPT-3.5 | | | GPT-4 | | |
|---|---|---|---|---|---|---|
|  | I/O* | COT# | P value | I/O | COT | P value |
| Osteoarthritis Benchmark | 0.16 | 0.17 | 0.41 | 0.24 | 0.23 | 0.52 |
| Guideline item QA (GIQA) | 0.26 | 0.28 | 0.03 | 0.38 | 0.38 | 0.80 |
| Management option QA (MOQA) | 0.22 | 0.20 | 0.004 | 0.30 | 0.27 | 0.002 |
| Treatment strategy QA (TSQA) | 0.02 | 0.03 | <0.001 | 0.07 | 0.07 | 0.79 |
| Real-case QA (RCQA) | 0.03 | 0.01 | <0.001 | 0.01 | 0.01 | 0.20 |

* Input Output prompt technique; # zero-shot chain of thought prompt technique.

**Human Evaluation Results**

From each of the GIQA, MOQA, TSQA, and RCQA, 20 items were randomly selected along with the corresponding responses generated by each model. A total of 1200 outputs were evaluated by physicians and patients. The results of the human evaluations of GPT-3.5, GPT-4, and DocOA revealed distinct outcomes across several aspects.

Figure 5a and Figure 5b shows the human evaluation results for the models' output. The rate of inaccuracy was the highest for GPT-3.5 (57%), followed by GPT-4 and DocOA at 50%, and 19.3%, respectively. All the models achieved high relevance and infrequently produced hallucinatory content in their responses. GPT-3.5 had a higher proportion of responses with missing content (22%) than GPT-4 (16.4%) or DocOA (16.5%). GPT-3.5 presented a higher likelihood of generating harmful content (20%) than GPT-4 (11.3%) and DocOA (8.3%). Moreover, GPT-3.5 was associated with a higher risk of causing severe harm (10.5%) than GPT-4 (5.5%) and DocOA (3.5%). The likelihoods of potentially biased content were 13.3%, 9.5%, and 2.8% for GPT-3.5, GPT-4, and DocOA, respectively. The results of the human evaluation for each subset benchmark are listed in Supplementary Files 2, 3, and 4. The results showed a substantial decrease in performance in terms of inaccurate content and missing content (GIQA > MOQA > TSQA > RCQA).

Figure 5. Results of Human Evaluation for the Assessment of Responses

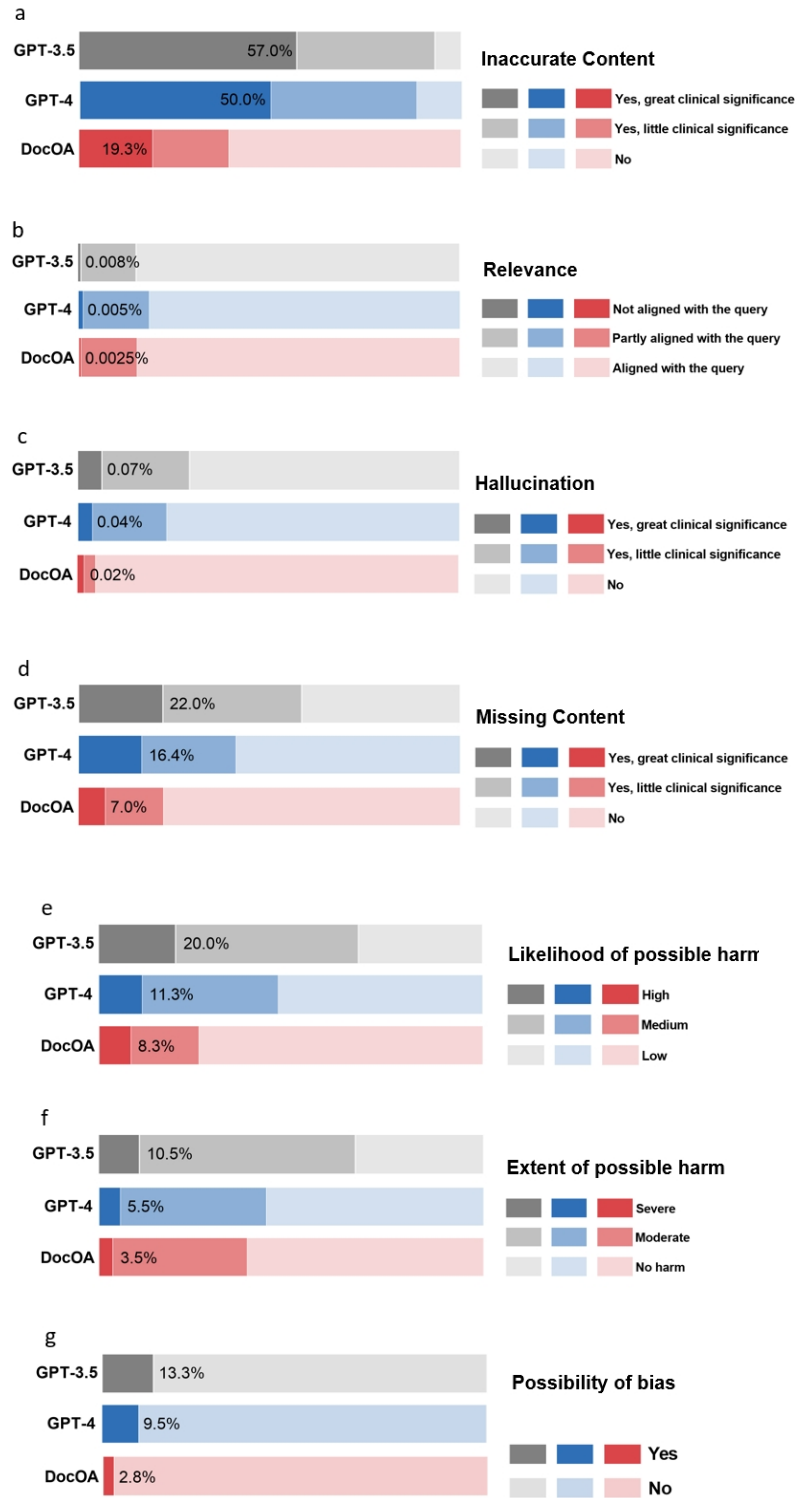

a. inaccurate content; b. relevance; c. hallucination; d. missing content; e. likelihood of possible harm; f. extent of possible harm; g. possibility of bias

Figure 6 shows the results of the LLMs' ability to assess correct comprehension, correct retrieval, and correct reasoning. Regarding the correct comprehension of the question, the response rate of DocOA was 91%, followed by GPT-4 (86%), and GPT-3.5 (82.5%). DocOA was able to correctly recall and present complete, relevant information in 65.8% of the responses, followed by GPT-4 (14.3%) and GPT-3.5 (12.0%). In terms of subset evaluation, the results showed comparable performance in comprehension and reasoning among the different models, whereas a substantial performance decrease was found in correct retrieval across GIQA, MOQA, TSQA, and RCQA.

Figure 6. Results of human evaluation for LLMs' comprehension, retrieval and reasoning ability

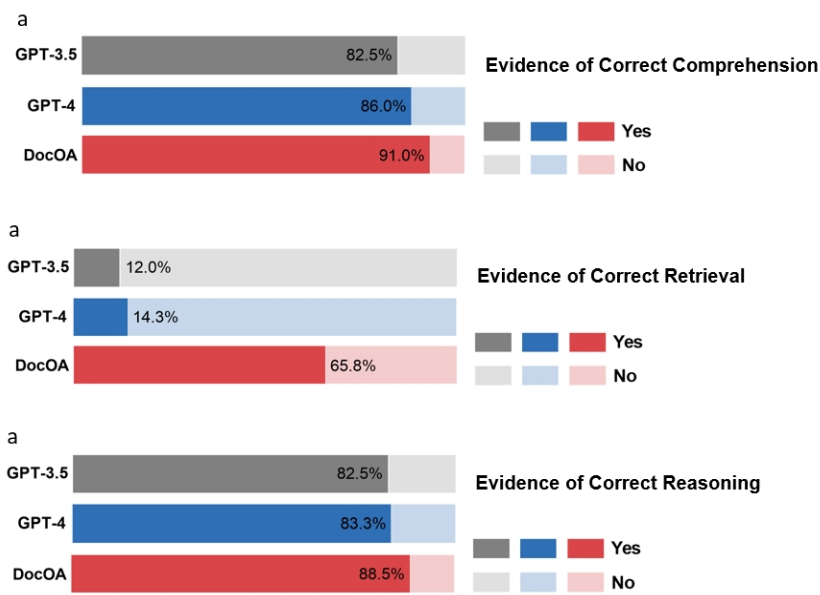

a. evidence of correct comprehension; b. evidence of correct retrieval; c. evidence of correct reasoning

The results of patient evaluations are shown in Figure 7. DocOA achieved a success rate of 71.3% in fulfilling patient intention, with GPT-4 at 39.8% and GPT-3.5 at 36.5%. Of the responses generated by DocOA, 75.8% were considered to be at least somewhat helpful, compared to 47% for GPT-3.5 and 47.75% for GPT-4. For GPT-3.5 and GPT-4, the subset evaluation showed a substantial decrease in intent fulfilment and helpfulness as the tasks shifted from domain-specific knowledge to personalised treatment recommendations.

Figure 7. Results of human evaluation by patients

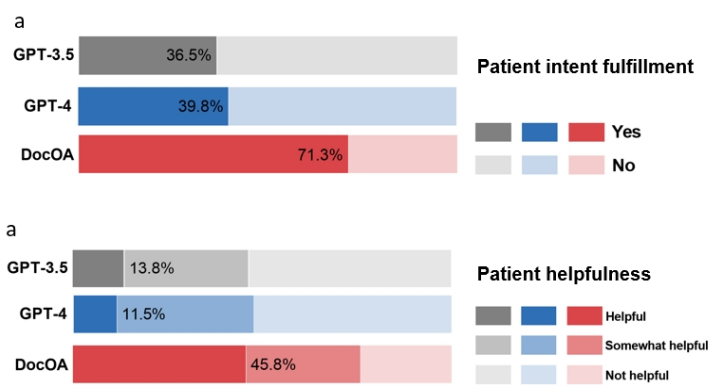

a. patient intent fulfillment; b. patient helpfulness

**Discussion**

This study introduced a benchmark framework to assess the performance of LLMs in specific medical domains. Using OA as a case study, this framework is the first to evaluate LLMs across a spectrum, from domain-specific knowledge to clinical applications in specific disease management. The incorporation of human evaluation provides multiple dimensions of assessment that are of considerable interest to clinical practitioners and patients, making it an essential tool for evaluating the clinical capabilities of LLMs.

The study findings revealed that general-purpose models such as GPT-3.5 and GPT-4 exhibit unsatisfactory performance when benchmarked against OA management. Additionally, both models demonstrated a marked decline in performance as tasks shifted from domain-specific knowledge to personalised treatment recommendations. Notably, the integration of RAG and instruction prompts substantially improved the domain-specific capabilities of general-purpose LLMs without additional training. The findings of this study demonstrated a cost-effective method for evaluating and enhancing the capabilities of LLMs in specialised medical fields.

**Domain-specific Medical Benchmark**

Although benchmarks targeting general medical knowledge have been previously developed[19-21], recent research has suggested that these benchmarks are only preliminary indicators of medical knowledge. The absence of tailored benchmarks in specific domains remains a potential challenge for evaluating the clinical effectiveness of LLMs[22,5]. Therefore, we developed a domain-specific benchmark focused on disease management for OA, which was selected for its prevalence, substantial disease burden, and complexity of its management strategies[6,7]. This benchmark was designed to test the domain-specific knowledge and clinical capabilities of LLMs. The benchmark comprised four parts, each testing the ability of LLMs at different levels, including the ability to provide evidence-based knowledge, summarising knowledge to formulate recommendations, providing management recommendations for different patient populations, and formulating personalised management plans for real-world patients. The benchmark was constructed based on established clinical guidelines and real-world patient information. Clinical guidelines offer comprehensive reviews of updated evidence and expert opinions, making them reliable sources of domain-specific medical knowledge. Through panel discussions involving physicians and data scientists, the questions were designed in a hybrid format, integrating both definitive and interpretative elements. Using this benchmark, we confirmed that general-purpose LLMs exhibit suboptimal performance in specialised domains. A significant performance gap was observed between domain-specific knowledge and clinical proficiency. This highlights the challenges faced by general-purpose LLMs in effectively applying specialised knowledge to clinical scenarios.

**Human Evaluation Framework**

Human evaluation is a crucial component in assessing the medical capabilities of large language models (LLMs) and offers a multidimensional assessment of their clinical capabilities. In this study, the human evaluation framework was modified based on a previous study, and hallucinations and relevance were added as additional criteria[3,5]. The evaluation criteria included accuracy, relevance, hallucinations, omissions, potential harm, and biased content. Moreover, the performance of LLMs in question comprehension, information retrieval, and medical reasoning was evaluated, as these are crucial abilities in tailoring patient-specific treatment. Patient evaluations primarily determine how responses address the user's intent and helpfulness. Although previous studies indicate a notable gap between objective benchmarking and human evaluation, our findings reveal a smaller discrepancy[3]. This could be attributed to the different knowledge domains and designs of the QA structure in this benchmark. The results of our study suggest that the GPT-3.5, GPT-4, and DocOA performed well in terms of hallucinations, comprehension, reasoning, and relevance. DocOA outperformed the other models in terms of accurate information, correct retrieval, and helpfulness as perceived by patients. This indicates that although generalised models are proficient in some areas, they remain inadequate in delivering the qualified responses required in a clinical context.

**Augmenting LLM with Domain-specific Ability**

Several techniques are available for developing medical LLMs, which primarily include integrating domain-specific knowledge during the training phase through techniques such as reinforcement learning with human feedback (RLHF)[23-25]. However, in this study, we focused on enhancing already-trained LLMs, such as GPT-4, by employing a suite of techniques, including RAG and instruction-based prompts. Similar methodologies have been applied to the development of specialised LLMs for chemical domain[26].

This approach was adopted for the following reasons: first, augmenting an existing model such as GPT-4 is more cost-effective than training a new model from scratch; second, advanced general-purpose models have been trained on diverse datasets, providing a broad base of general knowledge that can be beneficial for understanding and contextualising domain-specific information; third, techniques such as RAG and specialised prompting offer the convenience of being adjustable and refined over time, enabling easy adaptability to new evidence in the fast-evolving field of medicine.

Our results demonstrated that GPT-4 can effectively acquire domain-specific knowledge and clinical capabilities in the management of OA through a combination of approaches, including RAG and instruction prompts. This strategy can also be applied cost-effectively to other medical domains. Nonetheless, the efficacy of the RAG is contingent upon factors such as the size and quality of the data, retrieval techniques employed, and the underlying architecture of the LLM in use[27]. In situations that require processing large-scale data and addressing complex problems, a multiagent collaboration approach may be promising. In this case, each agent has different domain-specific abilities for achieving a more comprehensive solution.

**Limitations**

This study had several potential limitations that need to be addressed. First, OA management is highly complex, and our current dataset remains limited and requires continuous supplementation and updating. Therefore, establishing specialised groups dedicated to building and updating these LLM databases is imperative for diverse medical applications. Second, our reliance on English sources could restrict the applicability and inclusivity of our findings across different linguistic and cultural contexts. Third, although there have been human evaluations of clinical case data, the augmented model has not yet been tested in a real-world clinical setting. These limitations highlight the necessity for ongoing development and a comprehensive, multidimensional approach for evaluating LLMs in the medical field.

**Conclusion**

In this study, we introduce a novel benchmark framework designed to evaluate the capabilities of LLMs in specific medical domains, with OA serving as a case study. This framework assesses LLMs in terms of medical knowledge, evidence summarisation, and clinical capabilities. Through a combination of objective measures and human evaluations, we identified the limitations of generalised LLMs in clinical contexts. Furthermore, our study demonstrated that integrating RAG and instruction prompts significantly enhances the domain-specific performance of LLMs. This approach is a potentially cost-effective strategy for developing domain-specific medical LLMs.